\documentclass[letterpaper]{article} 
\usepackage[]{aaai25}  
\usepackage{helvet}  
\usepackage{courier}  
\usepackage[hyphens]{url}  
\usepackage{caption}
\usepackage{subfigure}
\usepackage{times}
\usepackage{helvet}
\usepackage{courier}
\usepackage[noend]{algpseudocode}
\usepackage{algorithmicx,algorithm}
\usepackage{multirow}
\usepackage{bibentry}
\usepackage{amsmath}
\usepackage{graphicx} 
\usepackage{natbib}  
\usepackage{caption} 
\usepackage{algorithmicx,algorithm}
\usepackage{algorithmicx}
\usepackage{algpseudocode}
\usepackage{multirow}
\usepackage{bm}  
\usepackage{colortbl}
\usepackage{booktabs}
\usepackage{newfloat}
\usepackage{listings}
\DeclareCaptionStyle{ruled}{labelfont=normalfont,labelsep=colon,strut=off} 
\floatstyle{ruled}
\newfloat{listing}{tb}{lst}{}
\floatname{listing}{Listing}
\title{Mjölnir: Breaking the Shield of Perturbation-Protected Gradients via Adaptive Diffusion}
\author{
    Xuan Liu\textsuperscript{\rm 1},
    Siqi Cai\textsuperscript{\rm 2},
    Qihua Zhou\textsuperscript{\rm 3},
    Song Guo\textsuperscript{\rm 4},
    Ruibin Li\textsuperscript{\rm 1},
    Kaiwei Lin\textsuperscript{\rm 2}
}
\affiliations{
    \textsuperscript{\rm 1}The Hong Kong Polytechnic University, Hong Kong, China\\
    \textsuperscript{\rm 2}Wuhan University of Technology, Wuhan, China\\
    \textsuperscript{\rm 3}Shenzhen University, Shenzhen, China\\
    \textsuperscript{\rm 4}The Hong Kong University of Science and Technology, Hong Kong, China\\
    xuan18.liu@connect.polyu.hk, 
    csiqi@whut.edu.cn, 
    qihuazhou@szu.edu.cn, songguo@cse.ust.hk\\
    ruibin.li@connect.polyu.hk, 
    297662@whut.edu.cn
}

\usepackage{bibentry}

\begin{document}

\maketitle

\begin{abstract}
Perturbation-based mechanisms, such as differential privacy, mitigate gradient leakage attacks by introducing noise into the gradients, thereby preventing attackers from reconstructing clients' private data from the leaked gradients. However, can gradient perturbation protection mechanisms truly defend against all gradient leakage attacks? In this paper, we present the first attempt to break the shield of gradient perturbation protection in Federated Learning for the extraction of private information. We focus on common noise distributions, specifically Gaussian and Laplace, and apply our approach to DNN and CNN models. We introduce Mjölnir, a perturbation-resilient gradient leakage attack that is capable of removing perturbations from gradients without requiring additional access to the original model structure or external data. 
Specifically, we leverage the inherent diffusion properties of gradient perturbation protection to develop a novel diffusion-based gradient denoising model for Mjölnir. By constructing a surrogate client model that captures the structure of perturbed gradients, we obtain crucial gradient data for training the diffusion model. We further utilize the insight that monitoring disturbance levels during the reverse diffusion process can enhance gradient denoising capabilities, allowing Mjölnir to generate gradients that closely approximate the original, unperturbed versions through adaptive sampling steps. 
Extensive experiments demonstrate that Mjölnir effectively recovers the protected gradients and exposes the Federated Learning process to the threat of gradient leakage, achieving superior performance in gradient denoising and private data recovery. 
\end{abstract}
\section{Introduction}
\begin{figure}[tb]
    \centering\includegraphics[width=0.47\textwidth]{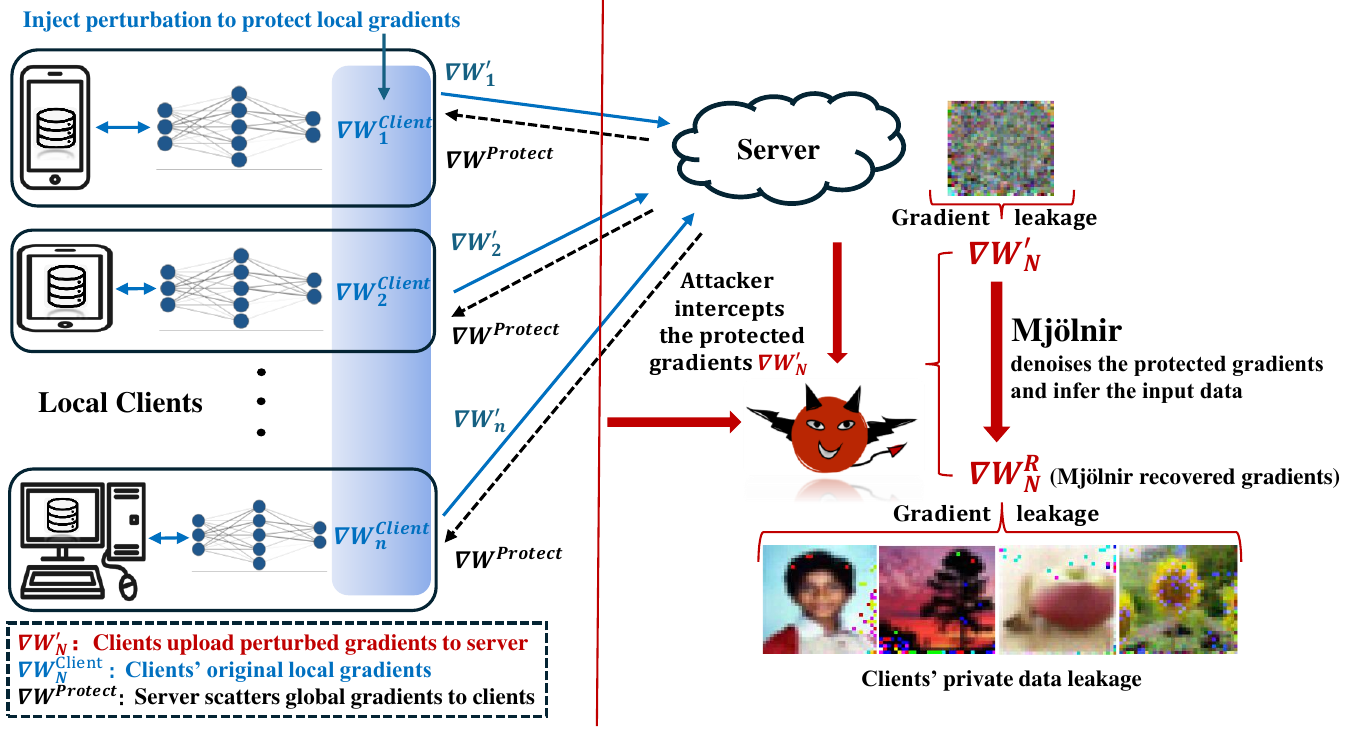}
    \caption{Threat model. The FL training process is threatened by gradient leakage attacks, where the attacker can intercept the exchanged gradients $\nabla W$ to recover the private training data. Previous work often protects the gradients by injecting perturbation into the gradients to form $\nabla W'_N$ and $\nabla W^{Protect}$. Our Mjölnir removes the perturbation injected in the protected gradients via the adaptive diffusion process.} 
    \label{fig:enter-label}
    \vspace{-12pt}
\end{figure}
\par Federated Learning (FL) is a distributed machine learning paradigm that facilitates collaborative model training without directly transmitting raw training data. Instead, it aggregates gradients or parameters shared among clients to build a global model~\cite{pmlr-v54-mcmahan17a, gong2024federated, 10092911, zhangFL}. This approach preserves the privacy of raw data by keeping it within its originating domain, addressing concerns associated with traditional centralized data processing. However, FL is vulnerable to gradient inversion attacks~\cite{10024757,Liu_Cai_Li_Zhang_Guo_2023, zhu2019deep}, in which adversaries can potentially reconstruct sensitive user data from the shared gradients. This vulnerability has spurred significant research into gradient protection techniques~\cite{tan2024defending, RODRIGUEZBARROSO2023148}. \textit{Gradient Perturbation}, such as differential privacy (DP), injecting noise into gradients to enhance privacy, has been proven to be an effective strategy for safeguarding data in FL scenarios~\cite{9069945,9714350,wang2024more}.

\par Perturbation-based gradient protection achieves its goal by adding noise to the gradients. If a method is developed to eliminate this noise effectively, the protective mechanism of this approach becomes ineffective. \textit{Diffusion Model}~\cite{NEURIPS2020_4c5bcfec,gong2023gradient} has the natural applicability to denoising the perturbation and is a potential approach to attack perturbation-based gradient protection. We hold the above idea based on two points: 1) Considering that gradient is the result of linear transformations applied to the training data, such as image data, we posit that if the original data are stable and predictable, then the corresponding gradient data will also be stable and predictable. 2) The essence of gradient perturbation protection is akin to the diffusion process applied to inherently structurally stable gradient data.

In this paper, we present the \textit{first} attempt to break the shield of gradient perturbation protection in FL based on the diffusion model. We reveal the natural diffusion properties of gradient perturbation protection and propose \textbf{Mjölnir}\footnote{Mjölnir is the hammer wielded by Thor and is renowned for its incredible power to break any shield.}, a perturbation-resilient gradient leakage attack method and the first general gradient diffusion attack (schematic diagram shown in Fig. 2) that is capable of rendering various kinds of perturbations on gradients (e.g. Differential Privacy \cite{10.1145/3378679.3394533}, certain layer representation perturbation \cite{Sun_2021_CVPR}, dynamic perturbation \cite{9546481}) nearly invalid. As the first attempt, we focus on common noise distributions, specifically Gaussian and Laplace, and apply our approach to DNNs and CNNs. Our method involves constructing a surrogate model for the target attack model to obtain protected shared gradients. In the absence of the original model structure and third-party datasets, we use the surrogate gradient data supply model to generate training data for our gradient diffusion model. This trained gradient diffusion model allows us to approximate the original shared gradients from the perturbed, privacy-preserving gradients obtained by attackers. Mjölnir integrates the perturbation protection mechanism into the reverse diffusion process, adjusting the privacy-preserving gradient denoising nodes and regulating the diffusion's forward and reverse time steps with the perturbation level $M$ as an adaptable parameter. 

In summary, our key contributions include:
\begin{itemize}
\item We disclose the natural diffusion process in general gradient perturbation mechanisms and introduce a perturbation adaptive parameter $M$ to adaptively adjust the diffusion step size according to the degree of perturbation.
\item We propose Mjölnir, the first practical gradient diffusion attack strategy to recover the perturbed gradients, without additional access to the original model structure and third-party data, which breaks the bottleneck that existing gradient leakage attacks cannot effectively leak privacy under gradient perturbation protection. 
\item We demonstrate the vulnerability of gradient perturbation protection under the Mjölnir adaptive diffusion denoising process. Experimental results under the general perturbation protection FL system show that Mjölnir achieves the best gradient denoising quality and privacy leakage ability on commonly used image datasets.
\end{itemize}

\section{Background and Related Work}
\subsection{FL with Perturbation Protection (FL-PP)}
FL-PP includes a variety of alternative gradient perturbation techniques, such as adding random global noise to gradients, noise to specific layers, and dynamic noise addition \cite{ijcai2022p791}. Among these, FL with Differential Privacy (FL-DP) is the most widely used method \cite{9912197,10.1109/TIFS.2023.3297369,tan2024defendingdatareconstructionattacks}. 
\par Generally, DP is defined based on the concept of adjacent database and has been applied in various practical areas in Artificial Intelligence to protect privacy information through adding specific perturbation, e.g. Google's RAPPORT \cite{10.1145/2660267.2660348} and large-scale graph data publishing \cite{8781882}. FL-DP protects the shared model parameters or gradients between clients and the server during the FL training process by applying Local Differential Privacy (LDP \cite{9253545, 10.1145/3378679.3394533}) and (or) Differential Privacy Stochastic Gradient Descent algorithm (DPSGD \cite{10.1145/2976749.2978318, 10008087}).
In this paper we discuss both $\varepsilon - DP$ and $(\varepsilon, \delta) - DP$ \cite{10.1007/11761679_29,10.1007/11681878_14,10.1145/2976749.2978318} as our sample attack background \cite{9912197,10.1109/TIFS.2023.3297369,tan2024defendingdatareconstructionattacks}:
\par \textbf{\textit{Assumption 1.}} A randomized mechanism $M$: $D \rightarrow R$ with domain $D$ and range $R$ satisfies $\varepsilon$ - differential privacy if for any two adjacent inputs $\textit{d}$, $\textit{d'}$ $\in D$ and for any subset of outputs $S \subseteq R$, and it holds that:
\begin{equation}
\label{Eq:def1}
Pr[M(d) \in S]\le e^\varepsilon Pr[M(d') \in S]
\end{equation}

\par \textbf{\textit{Assumption 2.}} A randomized mechanism $M$: $D \rightarrow R$ with domain $D$ and range $R$ satisfies $(\varepsilon, \delta)$ - differential privacy if for any two adjacent inputs $\textit{d}$, $\textit{d'}$ $\in D$ and for any subset of outputs $S \subseteq R$, and it holds that:
\begin{equation}
\label{Eq:def2}
Pr[M(d) \in S]\le e^\varepsilon Pr[M(d') \in S] + \delta
\end{equation}
$\delta$-approximation is preferably smaller than $1/|d|$.
We usually apply Laplace perturbation for the definition in Eq.~(\ref{Eq:def1}). However, as to the definition in Eq.~(\ref{Eq:def2}), the noise perturbation mechanism needs to be Gaussian Mechanisms, which can adapt both $\varepsilon$ and $\delta$. Refer to Theorem A.1 in Gaussian Mechanisms \cite{10.1561/0400000042}, with $\nabla s = max_{D, D'}||S(D)-S(D')||$ ($s$ is the real-value function) and $c_{dp}$ denotes the hyperparameter used to define the DP boundary, to ensure Gaussian noise distribution $N (0, \sigma^2_{dp})$ well preserves $(\varepsilon, \delta) - DP$, the noise scale should satisfy:
\begin{eqnarray}
\sigma_{dp} &\ge& c_{dp} \nabla s/\varepsilon ,\quad \varepsilon \in (0,1) \\
c_{dp} &\ge& \sqrt{2ln(1.25/\delta)}
\end{eqnarray}
\par Taking the local participates site for example, the main actions for FL-DP can be divided into four steps: setting DP noise mechanism, local gradient clipping, local gradient perturbation, and uploading the protected parameters to the server \cite{9069945}. FL involves multiple participants, making composition theorem in DP necessary when applying noise. Among the composition DP methods available, including Simple Composition, Advanced Composition \cite{10.1561/0400000042}, and Moments Accountant \cite{10.1145/2976749.2978318}, we utilize Simple Composition for the following discussion. In the case where $M_i$ satisfies $(\varepsilon, \delta)-DP$, the composition ($M_1, M_2, \cdots, M_k$) satisfies ($\sum_{i=1}^k\varepsilon_i, \sum_{i=1}^k\delta_i$). Aligning with the state-of-art FL-DP framework \cite{9069945,10.1007/11761679_29,10.1007/11681878_14,10.1145/2976749.2978318,9912197,10.1109/TIFS.2023.3297369,tan2024defendingdatareconstructionattacks}, We applied the following assumptions for noise calculation in local client gradient perturbation, considering both $\varepsilon$-DP (using the Laplace mechanism) and $(\varepsilon, \delta)$-DP (using the Gaussian mechanism):
\begin{eqnarray}
\textit{Laplace}: \sigma_{dpc} &=& \nabla s_{c} \times 1/\varepsilon\\
\label{Eq:def6}
\textit{Gaussian}: \sigma_{dpc} &=& \nabla s_{c} \times \sqrt{2ln(1.25/\delta)}/\varepsilon
\end{eqnarray}
where $\nabla s_{c}$ is the sensitivity and can be formulated as $\nabla s_{c} = \frac{2C}{m}$. $C$ is the clipping threshold for bounding $||\nabla W_i|| \le C$, where $\nabla W_i$ denotes the unperturbed gradient from the $i-th$ client and $m$ denotes the minimum size of local datasets.
\subsection{Gradient Inversion Attack (GradInv)}
\textit{GradInv} is a prevalent method in gradient leakage attacks and aims to steal the client's privacy information in the FL system. The primary idea for the majority of \textit{GradInv} to recover original information is minimizing the distance between the dummy gradient $\nabla W_{\xi}$ and the original gradient $\nabla W$ while updating the random data $x_{\xi}$ and label $y_{\xi}$ until the optimized results $\nabla W^*_{\xi}$ and $(x^*_{\xi},y^*_{\xi})$ are close enough to the original ones. The key formulation can be described as:
\begin{equation}
\quad min||\nabla W_{\xi}-\nabla W ||:(x_{\xi},y_{\xi})\rightarrow(x^*_{\xi}, y^*_{\xi})
\end{equation} 
\par Previous works utilize Peak Signal-to-Noise Ratio (PSNR) of the recovered images to evaluate the performance of \textit{GradInv}, with the threshold for the success of such attacks typically hinging on the degree of detail discernible to the human eye in the reconstructed private images \cite{geiping2020inverting,zhu2019deep,Liu_Cai_Li_Zhang_Guo_2023}.
\section{Methodology}
\begin{figure}[tb]
    \centering\includegraphics[width=0.47\textwidth]{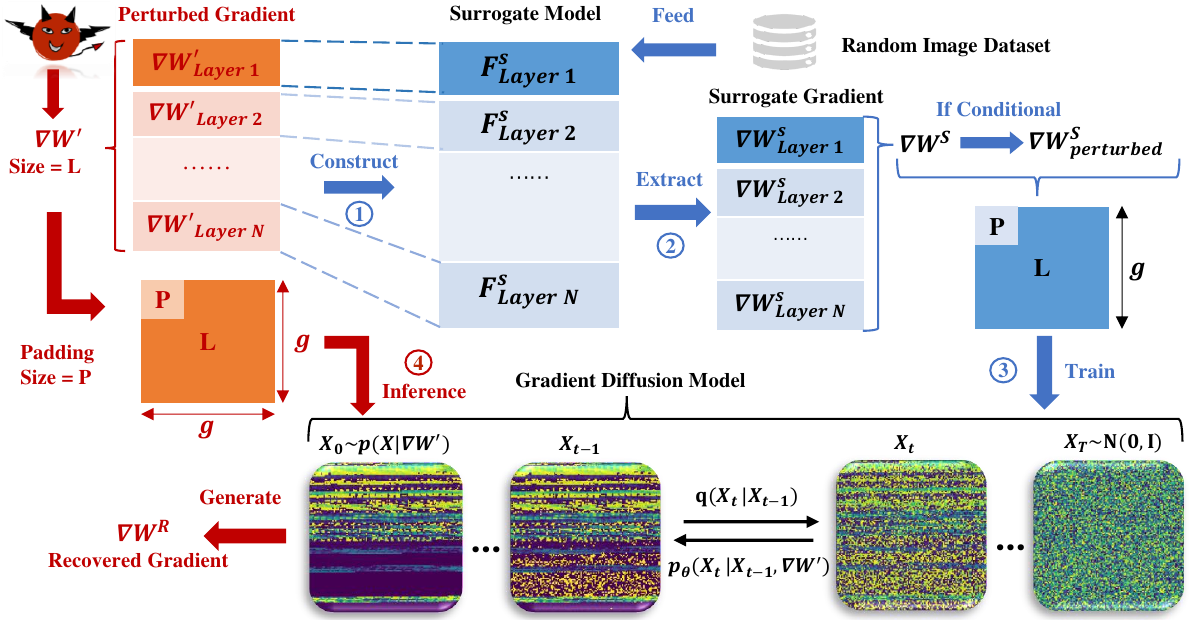}
    \caption{Mjölnir Overview. After intercepting the exchange gradients protected by unknown perturbation ($\nabla W'$) from clients, the attacker will (1) leverage the invariance of gradient data structure before and after perturbation to construct the surrogate model; (2) feed random image dataset into the surrogate model to extract surrogate clean gradients; (3) flatten the surrogate gradients and pad them to the appropriate size $g^2 = P+L$ (g is the minimum integer satisfies $g^2>L$) to create a surrogate gradient ($\nabla W^s$) dataset for training Gradient Diffusion Model (if additional conditions are chosen to guide the training process, joint ($\nabla W^s$,$\nabla W^s_{perturbed}$) or ($\nabla W^s$,$\nabla W'$) as the training dataset, where $\nabla W^s_{perturbed}$ denote surrogate gradients applied with known perturbation; if no conditions are needed, directly use $\nabla W^s$); (4) use the trained Gradient Diffusion Model to denoise $\nabla W'$ to generate the recovered gradient $\nabla W^R$. From $\nabla W^R$, the attacker can recover clients' privacy information.}
    \label{fig:enter-label}
    \vspace{-12pt}
\end{figure}
\par Mjölnir is the first Diffusion Attack Method focusing on the gradient data structure that can be applied to multiple kinds of gradient perturbation protection in FL through adaptive parameters setting on both forward and reverse process of the Gradient Diffusion Model employed in Mjölnir. The threat models targeted by Mjölnir encompass various forms of gradient perturbation. Meanwhile, the extremely similar noise mechanism on gradient perturbation protection and diffusion Markovian process provides a mathematical necessity for Mjölnir to realize an efficient privacy attack on gradient perturbation protection. Overall, our Mjölnir method (shown in Fig. 2) can be summarized in four steps:
\par \textbf{Step (1) Get Protected Gradients.} Honest-but-curious malicious attackers steal the Shared Perturbed Gradients $\nabla W'$ during the FL training process by hiding on the server side or waiting on the way of parameter sharing (Fig.1).

\begin{algorithm}
    \caption{Gradients Extracted for Mjölnir Training} 
    \small	
    \begin{algorithmic}[1]
        \State Stolen Protected Gradients: $F_i \rightarrow \nabla W'$;
        \State Construct surrogate Model: $\nabla W' \rightarrow F^s$;
        \State $j=0$;
        $ \epsilon \sim N(\textbf{0},\textbf{I})$;
        \For{$iteration=1$ to \textit{$Length_{random dataset}$}} 
        \State $\nabla W_j=\partial l(F^s(x_j;W^s),y_j)/\partial W^s$;
        \State \textbf{Save} $\nabla W_j$;
        \State $j = j+1$;
        \EndFor
    \end{algorithmic}
\end{algorithm}
\vspace{-12pt}
\begin{algorithm}
\caption{Gradient Diffusion Model Training} 
\small	
    \begin{algorithmic}[1]
        \If{With Condition $\nabla W'$}  $X_0 = (\nabla W', \nabla W_j)$;
        \Else  $ \quad X_0 = \nabla W_j $;
        \EndIf
        \State $\textbf{Repeat}$: 
        \State \quad $ X_0 \sim p(X_0)$;
        $ t \sim Uniform(1 \rightarrow T)$;
        $ \epsilon \sim N(\textbf{0},\textbf{I})$;
        \State \quad Take a gradient descent step on: 
    $\nabla_\theta||\epsilon - f_\theta (\sqrt{\gamma_t}X_0 + \sqrt{1-\gamma_t}\epsilon,t)||^2$;
    \State \textbf{until} converged;
    \end{algorithmic}
\end{algorithm}
\par \textbf{Step (2) Construct Surrogate Model.} Construct Surrogate Gradients Data Supply Model $F^s$ from the data structure of Shared Perturbed Gradients $\nabla W'$ stolen by the attacker that can output the same gradient data structure as the target attacked client's local model (Algorithm 1, Fig. 2). 
 \par Before delving into Step (3) and Step (4), two configurations that recur in the subsequent steps are defined:
\par \textit{\textbf{[Gradient Adjustment]}}: Gradients each with total size $L$ ($L = L_{\nabla W_{Layer 1}} + L_{\nabla W_{Layer 2}}+...+ L_{\nabla W_{Layer N}}$) are adjusted into $1 \times g \times g$ ($g^2 = L+P$: $g=$ minimum integer satisfies $g^2 > L$; $P=$ 0-padding size) before feeding into Gradient Diffusion Model for training or inference to adapt gradient diffusion process. The adjustment is only related to Diffusion procedures, gradients are restored to their original structure and size before further Gradient-Based Attacks and Evaluations.
\par \textit{\textbf{[M-Adaptive Process]}}: Adaptive parameter $M$ is inserted into $\nabla W'$ before inference to ensure the starting time step is appropriately positioned to maximize the denoising capability of Gradient Diffusion Model (Eq.~(\ref{Eq:def16}) $\sim$ Eq.~(\ref{Eq:def18})).
\par \textbf{Step (3) Train Gradients Diffusion Model.} Construct our Gradient Diffusion Model that takes into account the level of knowledge about the attacked model (e.g., whether the level of perturbation noise or the type of noise is known). Conduct \textit{[Gradient Adjustment]} to Clean Gradients $\nabla W^s$ extracted from Surrogate Model $F^s$ to build a training gradient dataset to train our Gradient Diffusion Model (Fig. 2, Fig.3). Algorithm 2 demonstrates a detailed training process.

\par \textbf{Step (4) Recover Original Gradients.} 
The stolen Shared Perturbed Gradients $\nabla W'$ are put into our trained Mjölnir Gradient Diffusion Model after \textit{[Gradient Adjustment]} and \textit{[M-Adaptive Process]} to generate the Recovered Gradients $\nabla W^R$ for further Gradient-Based Attack to get certain privacy information based on various demands (Algorithm 3, Fig. 2 and Fig. 3). 

\begin{figure}
    \centering\includegraphics[width=0.47\textwidth]{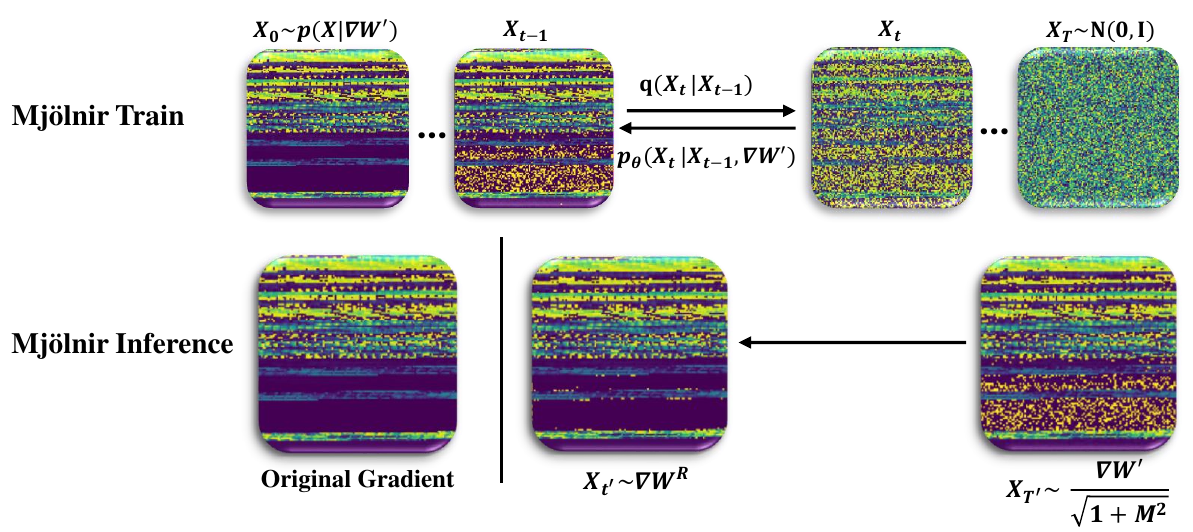}
    \caption{Visualization of Markovian gradient diffusion process. $M$ is the noise scale of perturbation, which is set as the adaptive parameter in \textit{[M-Adaptive Process]}. Mjölnir Train and Mjölnir Inference correspond to Algorithm 2 and Algorithm 3 respectively. }
    \label{fig:enter-label}
    \vspace{-12pt}
\end{figure}
\begin{algorithm}
\caption{Generate Original Gradient} 
\small	
\begin{algorithmic}[1]
\If{Known Noise Scale $ M $}  $ \textbf{c} = \frac{1}{\sqrt{1+M^2}}$;
\Else  $\quad \textbf{c} \in (0,1) $;
\EndIf
\State $X_{T'} \sim \textbf{c} \nabla W'$;
\For{$t = T',...,1$}
    \If{$t > 1$} $ \textbf{z} \sim N(\textbf{0},\textbf{I})$;
    \Else $ \quad \textbf{z} = \textbf{0}$;
    \EndIf
\State $X_{t-1} = \frac{1}{\sqrt{\alpha_t}}(X_t - \frac{1-\alpha_t}{\sqrt{1-\gamma_t}}f_\theta(X_t,t)) + \sqrt{1-\alpha_t}\textbf{z}$;
\EndFor
\State \Return $X_0$;
$\nabla W^R \leftarrow X_0$: \Comment{For Further GradInv};
\end{algorithmic}
\end{algorithm}
\par Mjölnir's Gradient Diffusion Model is inspired by DDPM\cite{NEURIPS2020_4c5bcfec}, to be specific:
\par In \textbf{Step (3) Train Gradients Diffusion Model}, refer to Algorithm 2 lines 4 to 6, we set $t \in (1, T)$ as the time steps of Gaussian noise addition. $T$ is the total time step of the forward Markovian diffusion process. $\alpha_t (0< \alpha_t < 1)$ denote the adaptive variables at each iteration and $\gamma_t = \prod_{t=0}^t \alpha_t$. With the above settings, the forward process ($q$: with no learnable parameters) of Mjölnir's Gradients Diffusion Model can be modeled as:
\begin{eqnarray}
\label{Eq:def10}
q(X_1|X_{0}) &=& N(X_1; \sqrt{\alpha_1}X_{0}, (1-\alpha_1)I) \\
\label{Eq:def11}
q(X_{t}|X_{0}) &=& N(X_t; \sqrt{\gamma_t}X_0,(1-\gamma_t)I) 
\end{eqnarray}
Given $(X_0,X_t)$, $X_{t-1}$ can be modeled as:
\begin{equation}
\label{Eq:def12}
q(X_{t-1}|X_{0},X_t) = N(X_t; \mu_t,\sigma_tI)
\end{equation}
\par If insert the stolen Shared Perturbed Gradients $\nabla W'$ or constructed Surrogate Perturbed Gradients $\nabla W^s_{perturbed}$ as training condition then input $X_0=(condition,\nabla W_j)$.If not to train with the condition, then input $X_0=\nabla W_j$.Through algebraic calculation, $\mu_t$ and $\sigma_t$ in Eq.~(\ref{Eq:def12}) can be simplified for further usage in the reverse training process\cite{9887996} as:
\begin{eqnarray}
 \mu_t &=& \frac{\sqrt{\gamma_{t-1}}(1-\alpha_t)}{1-\gamma_t}X_0 + \frac{\alpha_t(1-\gamma_{t-1})}{1-\gamma_t}X_t \\
\sigma^2_t &=& \frac{1-\gamma_{t-1}}{1-\gamma_t}(1-\alpha_t)
\end{eqnarray}
\par In the reverse process $(p)$ of Mjölnir's Gradient Diffusion training, the objective function $f_\theta$ which is trained to predict noise vector $\epsilon$ is modeled as: 
\begin{equation}
E_{X_0,\epsilon}[\frac{(1-\alpha_t)^2}{2\sigma^2_t\alpha_t(1-\gamma_t)}||f_\theta (\sqrt{\gamma_t}X_0 + \sqrt{1-\gamma_t}\epsilon,t) -\epsilon||^2]
\end{equation}
\begin{table*}[htbp]
\centering
\setlength{\tabcolsep}{1.8mm}{
\begin{tabular}{c|c|cc|cc|cc}
\hline

\hline
\multirow{2}{*}{\textbf{Datasets}} & \multirow{2}{*}{\textbf{Model}} & \multicolumn{2}{|c|}{\textbf{$\bm{\varepsilon}$=1}} & \multicolumn{2}{c|}{\textbf{$\bm{\varepsilon}$=5}} & \multicolumn{2}{c}{\textbf{$\bm{\varepsilon}$=10}} \\ \cline{3-8} 
                          &                        & $PSNR_{i}$            & $LRA$              & $PSNR_{i}$           & $LRA$              & $PSNR_{i}$            & $LRA$               \\ \hline
\multirow{6}{*}{MINIST}  & GRNN\cite{ren2022grnn}                   & 14.70             & 1.000                & 16.54             & 1.000                & 20.19             & 1.000                 \\
                          & IG\cite{geiping2020inverting}     & 3.721                 & -                & 5.164                & -                & 6.067                & -                 \\
                          & DLG\cite{zhu2019deep}                    & 0.222               & 0.676            & 8.025               & 0.752            & 17.08               & 0.909             \\
                          
                          & Mjölnir            & 17.87  \cellcolor[gray]{0.85}                 & 0.851               & 24.17  \cellcolor[gray]{0.85}                 & 0.895               & 33.09   \cellcolor[gray]{0.85}                & 0.927                 \\
                          
                         & Mjölnir(Non-Adaptive) & 17.39 \cellcolor[gray]{0.85}                  & 0.851              & 24.14  \cellcolor[gray]{0.85}                 & 0.889             & 32.98    \cellcolor[gray]{0.85}               & 0.925                 \\
                          & Mjölnir(Conditional)  & 20.87 \cellcolor[gray]{0.85}                  & 0.887               & 28.39    \cellcolor[gray]{0.85}               & 0.924              & 35.14    \cellcolor[gray]{0.85}               & 0.957                \\ \hline
\multirow{6}{*}{CIFAR100} & GRNN\cite{ren2022grnn}                   & 18.24              & 1.000                & 20.61              & 1.000                & 21.15              & 1.000                 \\
                          & IG\cite{geiping2020inverting}     & 3.684                & -                & 5.073                & -                & 5.971                & -                 \\
                          & DLG\cite{zhu2019deep}                    & 0.112               & 0.769            & 7.673               & 0.833           & 18.57             & 0.896             \\
                          & Mjölnir                     & 18.25 \cellcolor[gray]{0.85}                   & 0.883     & 19.04                  & 0.902                 & 21.32 \cellcolor[gray]{0.85}                & 0.902              \\
                          & Mjölnir(Non-Adaptive)      & 18.15                   & 0.871               & 17.98                & 0.884                & 21.06                  & 0.891                 \\
                          & Mjölnir(Conditional)       & 18.69 \cellcolor[gray]{0.85}                    & 0.901          & 21.02 \cellcolor[gray]{0.85}                 & 0.902                & 23.49 \cellcolor[gray]{0.85}                 & 0.908                 \\ \hline
\multirow{6}{*}{STL10}    & GRNN\cite{ren2022grnn}                   & 12.24               & 1.000                & 16.24              & 1.000                & 20.98              & 1.000                 \\
                          & IG\cite{geiping2020inverting}     & 4.323                & -                & 6.082                & -                & 7.851                & -                 \\
                          & DLG\cite{zhu2019deep}                    & 0.038               & 0.783            & 5.673               & 0.815            & 17.25             & 0.851             \\
                          & Mjölnir                   & 12.36 \cellcolor[gray]{0.85}           & 0.806              & 16.30 \cellcolor[gray]{0.85}               & 0.857               & 19.77             & 0.889              \\
                          & Mjölnir(Non-Adaptive)  & 12.28  \cellcolor[gray]{0.85}                & 0.801               & 15.19                  & 0.849              & 19.76            & 0.887             \\
                          & Mjölnir(Conditional)     & 12.98 \cellcolor[gray]{0.85}                 & 0.816               & 20.30 \cellcolor[gray]{0.85}                & 0.872             & 22.77 \cellcolor[gray]{0.85}               & 0.906               \\ 
                          \hline

                          \hline

\end{tabular}}
\caption{Privacy leakage capability of Mjölnir variant models and traditional Gradients Leakage Attacks in FL-DP ($\delta$ = $10^{-5}$, $\varepsilon$ = 1, 5, 10). Gray marker: Mjölnir outperforms the highest result of traditional ones. }
\vspace{-12pt}
\end{table*}
\par In \textbf{Step (4) Recover Original Gradients}, refer to Algorithm 3, two different original gradient generation processes are chosen depending on whether or not attackers know the Noise Scale of Perturbation of the Threat Model. Take FL with Gaussian Differential Privacy (Eq.~(\ref{Eq:def6})) as an example, consider an experienced and clever attacker who may know the DP Privacy Budget $\varepsilon$ and the probability of information leakage $\delta$ ($\delta$ is likely to be set as $10^{-5}$ from usual practice in DP), combining with the sensitivity $\nabla S$ (which can be estimated if the attacker has some previous information with the target model training dataset), the noise scale can be calculated or estimated to an approximate value $M$  ($M$ defined as the adaptive parameter in \textit{[M-Adaptive Process]}). So, $\nabla W'$ can be modeled as:
\begin{equation}
\label{Eq:def16}
\nabla W' = \nabla W+ MN(0,I)
\end{equation}
\par Considering the forward Markovian gradient diffusion process in Step 3 Eq.~(\ref{Eq:def10}) $\&$ Eq.~(\ref{Eq:def11}), the relation between stolen perturbed gradients $\nabla W'$ and original gradients $\nabla W$ can be remodeled as:
\begin{equation}
\label{Eq:def17}
\frac{1}{\sqrt{1+M^2}} \nabla W' = \frac{1}{\sqrt{1+M^2}} \nabla W + \frac{M}{\sqrt{1+M^2}}N(0,I)
\end{equation}\\
Recall that our target is to generate $\nabla W^R$, so $\frac{1}{\sqrt{1+M^2}} \nabla W'$ should be considered as $X_t$ in the inverse process of Gradient Diffusion Model, while $\nabla W$ stands for $X_0$. 
\par Correspondingly, during the construction of the recovered gradient $\nabla W^R$, the inverse time steps can be set as $T'$. The relationship between $T'$ and $M$ is:
\begin{equation}
\label{Eq:def18}
\frac{1}{1+M^2} = \prod_{t=0}^{T'} \alpha_t
\end{equation}
\par Since $\gamma_t = \prod_{t=0}^t \alpha_t$ have been predefined and calculated during the forward Markovian diffusion process, $T'$ can be fixed in an approximate range $T' \in (T'_-, T'_+)$, where $\prod_{t=0}^{T'_-}\alpha_t < \frac{1}{1+M^2} < \prod_{t=0}^{T'_+}\alpha_t$ and $T'_{-or+}$ are positive integers less than the total forward noise addition time step $T$.
On the other hand, if $M$ can not be estimated, which means the attacker knows nothing about the noise scale. Since $\frac{1}{1+M^2}\in (0,1)$, the input value of the inference process can simply be set as $c\nabla W'$ where $c\in (0,1)$ to adapt the Markovian forward process.
\par Also, if conditions allow, $c$ can be modeled and predicted by a separate Machine Learning model according to the specific requirement of attackers.
\begin{figure}[]
    \centering
    \includegraphics[width=0.47\textwidth]{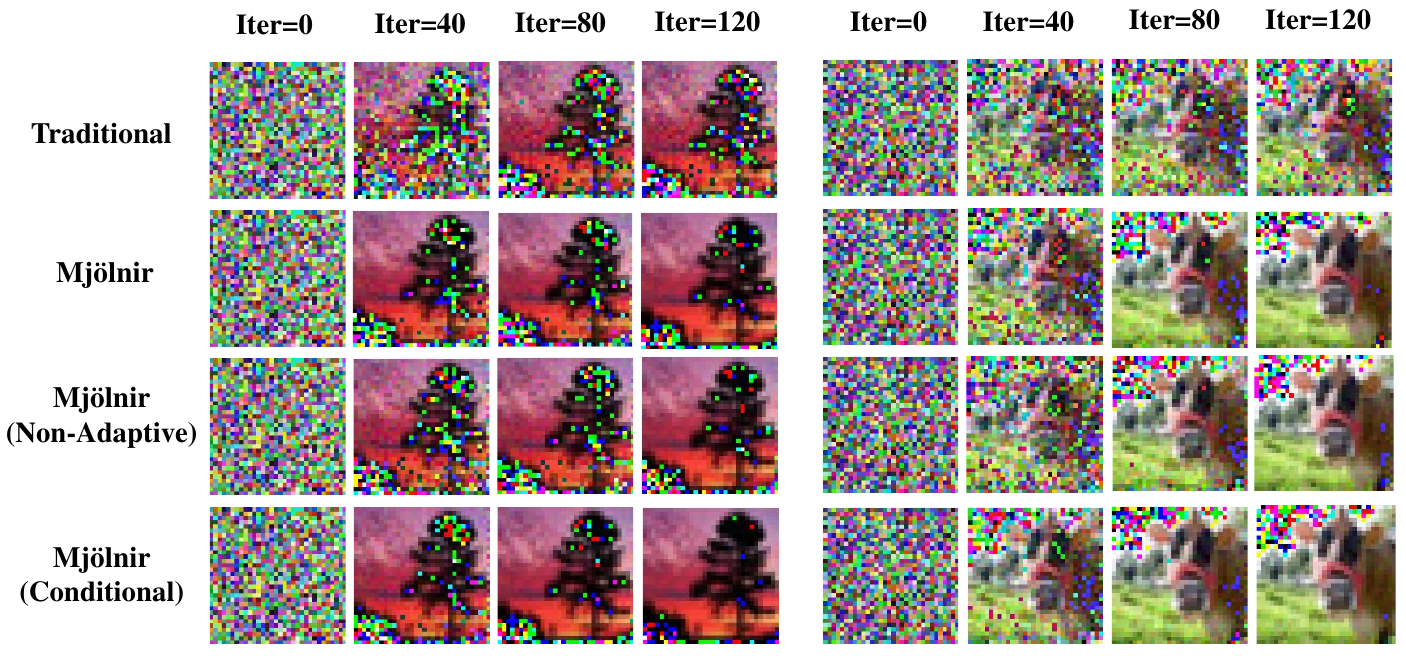}
    \caption{Comparison of the private image recovery procedures to the iterations between Mjölnir variant models and traditional Gradient Leakage Attack methods (DLG~\cite{zhu2019deep}).}
    \vspace{-12pt}
\end{figure}
 \begin{figure*}[htbp]
    \centering
    \includegraphics[width=0.75\textwidth]{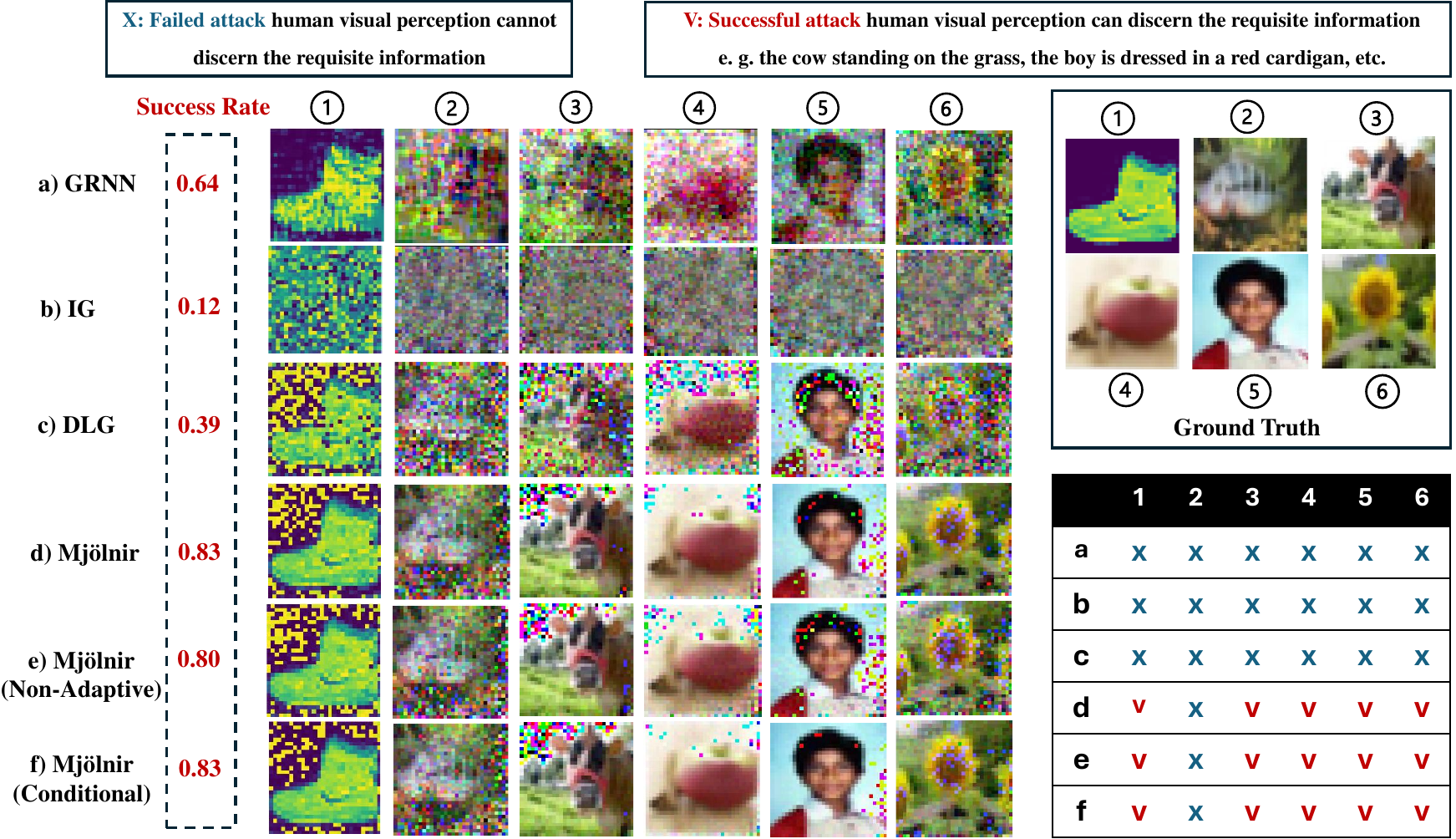}
    \caption{Comparisons on ground truth clients' private images and corresponding recovered privacy images from Mjölnir variant models and commonly used traditional gradient leakage attacks. ($\delta$ = $10^{-5}$; $\varepsilon=10$; Success Rate: overall attack success rate)}
    \vspace{-12pt}
\end{figure*}
\section{Experiments}
\subsection{Experimental Setups}
\textbf{(A) Mjölnir Variant Attack Models.} Mjölnir (trained with only unperturbed surrogate gradients), Conditional Mjölnir (trained with both perturbed gradients and unperturbed surrogate gradients), and Non-Adaptive Mjölnir (without \textit{[M-Adaptive Process]}: no perturbation scale $M$ as an adaptive parameter during gradient diffusion process) are presented. \\
\textbf{(B) Benchmarks and Datasets.} We employ MNIST, CIFAR100, and STL10 as client privacy datasets, which also serve as the ground truth for privacy leakage evaluation. We extract the unperturbed original gradients ($\nabla W$) of the aforementioned three datasets from the local training model of the target client as the reference benchmark of gradient denoising under the FL-PP paradigm. The Mjölnir gradient diffusion model is trained with gradients extracted from a separate dataset, FashionMNIST.\\
\textbf{(C) Evaluation and Boundary.} To evaluate the Privacy Leakage Capability, we utilize the Image Average Peak Signal-to-Noise Ratio $PSNR_{i}$ and the Label Recovered Accuracy $LRA$. These metrics are employed to assess the fidelity of the recovered images and the accuracy of the recovered labels, respectively, to the original images and ground truth labels. The boundary of Privacy Leakage Attack is aligned with previous works: the attack is considered successful if human visual perception can discern the requisite information from the recovered images. In the context of evaluating Gradient Denoising, we employ the cosine similarity $CosSimilar_{g}$ and the Average Peak Signal-to-Noise Ratio $PSNR_{g}$ as metrics to assess the quality of the Recovered Gradients $\nabla W^R$ compared to the Original Gradients $\nabla W$. Higher values of the two metrics indicate better accuracy in the original gradient recovery.
\subsection{Privacy Leakage Capability}
\par The comparison of overall privacy leakage capability from perturbed gradients of Mjölnir and traditional Gradient Leakage Attacks (GRNN \cite{ren2022grnn}, IG \cite{geiping2020inverting}, and DLG \cite{zhu2019deep}) under FL-DP, compares Image Average Peak Signal Noise Ratio $PSNR_{i}$ and Label Recovered Accuracy $LRA$ of the recovered local clients' privacy information. Local clients' private training datasets are MNIST, CIFAR100, and STL10.
\par According to the numerical experimental results shown in Table 1 and the visualization results shown in Fig. 5, Mjölnir variant models exhibit a substantial superiority over traditional gradient leakage attacks in terms of private image leakage. On average, there is an approximately $209\%$ increase in the recovered image $PSNR_{i}$ when using Mjölnir. For traditional methods, GRNN consistently achieves an LRA metric of 1.000 by using a robust generative network for label recovery, while other methods update images and labels jointly, leading to less accurate results. Moreover, upon examining the private image recovery procedures of Mjölnir variant models and traditional gradient leakage attacks illustrated in Fig. 4, it becomes evident that the attacks incorporating Mjölnir not only achieve considerably improved accuracy in the ultimate reconstruction of private images compared to conventional approaches but also showcase notable advantages in terms of attack iteration rounds and speed (Mjölnir (Conditional) $>$ Mjölnir $>$ Mjölnir (Non-Adaptive)). Above all, Mjölnir variant models demonstrate significant advantages in recovering private images.
\begin{table*}[htbp]
\centering

\label{tab:results}

\setlength{\tabcolsep}{3.5pt} 
\begin{tabular}{@{}c|cc|cc|cc@{}}
\toprule
\textbf{Model} & \multicolumn{2}{c|}{\textbf{MNIST}} & \multicolumn{2}{c|}{\textbf{CIRAF100}} & \multicolumn{2}{c}{\textbf{STL10}} \\
\cmidrule(lr){2-3} \cmidrule(lr){4-5} \cmidrule(lr){6-7}
& $CosSimilar_{g}$ & $PSNR_g$ & $CosSimilar_{g}$ & $PSNR_g$ & $CosSimilar_{g}$ & $PSNR_g$ \\
\midrule
NBNet\cite{Cheng_2021_CVPR} & 0.992 & 35.74 & 0.979 & 27.51 & 0.979 & 27.23 \\
SS-BSN\cite{han2023ssbsn} & 0.995 & 32.35 & 0.845 & 22.85 & 0.845 & 22.84 \\
AP-BSN\cite{9878719} & 0.968 & 29.61 & 0.892 & 24.01 & 0.893 & 24.12 \\
\hline
Mjölnir & 0.996 & 38.76 & 0.990 & 30.42 & 0.990 & 30.01 \\

Mjölnir(Non-Adaptive) & 0.995 & 38.59 & 0.990 & 30.39 & 0.990 & 29.87 \\

Mjölnir(Conditional)& 0.996 & 38.78 & 0.990 & 30.53 & 0.993 & 30.22 \\
\bottomrule
\end{tabular}
\caption{Overall results on gradients denoising in FL-DP. Threat model setting: DP-($\varepsilon$, $\delta$) = (2,$10^{-5}$) with Gaussian perturbation.}
\vspace{-12pt}
\end{table*}

\begin{figure*}[tb]
    \centering
    \includegraphics[width=0.91\textwidth,height = 0.25\textwidth]{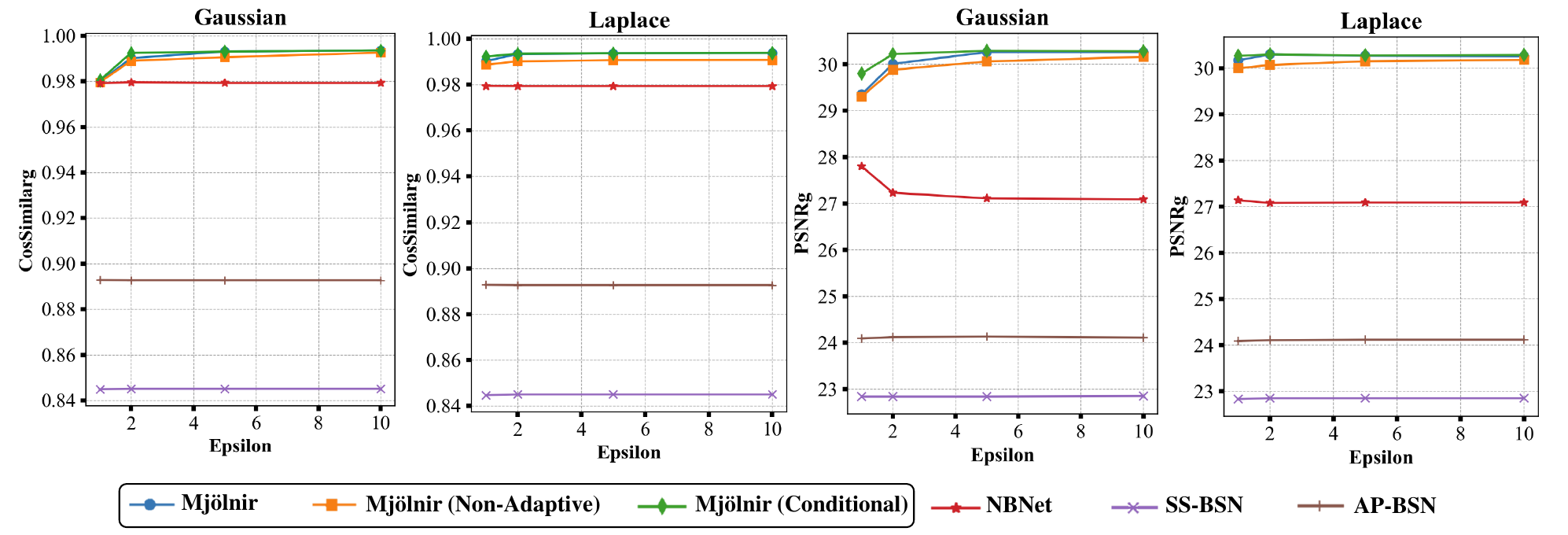}
    \caption{Gradients denoising under FL-PP: Gaussian and Laplace perturbed gradients denoising performance of $PSNR_{g}$ and $CosSimilar_{g}$ via different perturbation magnitudes. The same noise magnitude parameter $\varepsilon$ as in FL-DP is used to represent different levels of perturbation. Perturbation decreases, $\varepsilon$ increases. (Private dataset: STL10, $\delta$ = $10^{-5}$)}
    \vspace{-12pt}
\end{figure*}
\subsection{Gradients Denoising under FL-DP}
\par Among FL-DP, we choose the NbAFL framework \cite{9069945} (Noising before model aggregation FL) as the threat model in this experiment due to its widespread adoption. To ensure a comprehensive evaluation of the effectiveness of the Mjölnir, we train the three Mjölnir variant models, as well as non-diffusion denoising models such as NBNet \cite{Cheng_2021_CVPR}, SS-BSN \cite{han2023ssbsn}, and AP-BSN \cite{9878719}, using gradients extracted from the surrogate downstream task model trained on the FashionMnist. Following that, we utilize the trained models to denoise the shared perturbed gradients intercepted by the attacker from the target clients. These target clients have locally trained downstream task models using privacy datasets (MNIST, CIFAR100, and STL10). The gradients are protected using the NbAFL\cite{9069945} before being shared with the server.
\par Based on the experimental results presented in Table 2, it is evident that Mjölnir showcases superior denoising ability for perturbed gradients under FL-DP, with an average Cosine Similarity exceeding 0.992 and a PSNR of 37.68. This represents a significant improvement of over $27\% $ compared to non-diffusion methods. Further analysis reveals that among the three Mjölnir variant models, the performance can be ranked as follows: Mjölnir(Conditional)$>$ Mjölnir $>$ Mjölnir (Non-Adaptive). This ranking demonstrates both \textit{M-Adaptive Process} and Conditionalal training enhance the denoising and generation performance of Mjölnir's Gradient Diffusion Model. It suggests that selecting an appropriate Mjölnir variant model based on available prior information can improve attack performance.
\begin{table}[]
\centering

\begin{tabular}{c|c}
\hline
{\textbf{Model}} & {\textbf{Inference time (s)}} \\
\hline
NBNet\cite{Cheng_2021_CVPR}     &    2.653              \\
SS-BSN\cite{han2023ssbsn}    &    26.15                             \\
AP-BSN\cite{9878719}    &   7.138                              \\\hline
Mjölnir      &  6.834         \\
Mjölnir(Non-Adaptive)  & 6.834                \\
Mjölnir(Conditional)      & 6.917                      \\ 
\hline
\end{tabular}
 \caption{Gradient denoising average inference time of Mjölnir variant models and non-diffusion denoising models under FL-PP. (Private dataset: STL10, $\varepsilon$ = 0.5 $\sim$ 10, $\delta$ = $10^{-5}$. Device: NVIDIA GeForce RTX 2060 GPU; Intel(R) Core(TM) i7-10870H CPU at 2.20GHz.)}
 \vspace{-20pt}
\end{table}

\subsection{Gradients Denoising under FL-PP}
\par To evaluate the generalization capability of Mjölnir on gradient denoising, we constructed two different types of noise (Laplace and Gaussian) randomly applied to each layer of the original gradients within the FL-PP framework. The training and inference processes of Mjölnir variant models, and non-diffusion denoising models (NBNet \cite{Cheng_2021_CVPR}, SS-BSN \cite{han2023ssbsn}, and AP-BSN \cite{9878719}) are the same as the above experiments on gradients denoising under FL-DP. In contrast to attacks specifically tailored for the FL-DP framework, the gradient denoising experiments on FL-PP showcase Mjölnir's ability to effectively handle various types of perturbations and adapt to different magnitudes of perturbation.
\par Referring to the experimental results illustrated in Fig. 6, it is observed that when subjected to Laplace perturbation, Mjölnir variant models exhibit an average improvement of $21.6\%$ in PSNR and $9.2\%$ in Cosine Similarity. Similarly, under Gaussian perturbation, Mjölnir achieves an average enhancement of $21.3\%$ in PSNR and $9.1\%$ in Cosine Similarity. These findings provide compelling evidence for the robustness and stability of the Mjölnir variant models.
\par We further compared the average inference time of different denoising models under FL-PP. The experimental results presented in Table 3 indicate that Mjölnir variant models exhibit a relative advantage in terms of gradient denoising speed, surpassing non-diffusion methods by an average improvement of $32.7\%$ in inference time. \\
\textbf{Limitations: } (1) Mjölnir is not effective in attacking perturbations that are not based on gradient diffusion (noise perturbation) such as representation perturbation. (2) The overall effectiveness of Mjölnir in reconstructing privacy datasets is also bounded by the selected subsequent Gradient Leakage Attacks. (3) Mjölnir is used to attack CNN and (or) DNN-based models. Experimental results suggest that it struggles with transformer-based models due to their complex structure and larger number of parameters, making it difficult to accurately denoise gradients and recover data. 
\\
\textbf{Defense Strategies:} Possible defense strategies against Mjölnir can be approached by preserving the privacy of original data, preserving the target attack models, and shared gradients protection. Mjölnir effectively leaks privacy data by attacking the perturbed shared gradients, which suggests that the possible defense approaches against Mjölnir should either be non-perturbation gradient protection methods or non-gradient privacy-preserving methods. For instance, processing of the participant's local training model in depth and layers to weaken the performance of Gradient Leakage Attacks, encryption on original local privacy data before training, and various cryptography-based protections on shared gradients \cite{huang2021evaluating,ijcai2022p791, RODRIGUEZBARROSO2023148}. Cryptography-based approaches always cause more computation time and storage \cite{254465}, thus lightweight defense strategies that can serve as substitutes for invalid perturbation protection still need future explorations.

\vspace{-5pt}
\section{Conclusion}
This paper makes the first attempt to investigate the diffusion property of the widely used perturbation-based gradient protection. To reveal potential vulnerabilities, we propose a novel Perturbation-Resilient Gradient Leakage Attack via an adaptive diffusion process. This effective attack paradigm deactivates perturbation protection by leveraging the denoising capability of diffusion models without access to clients' models and external data. Based on Mjölnir, we wish to enhance public consciousness of the privacy leakage issues of existing perturbation-based defenses on gradients.

\bibliography{aaai25}
\end{document}